\documentclass[conference]{IEEEtran}
\IEEEoverridecommandlockouts
\usepackage{cite}
\usepackage{amsmath,amssymb,amsfonts}
\usepackage{algorithmic}
\usepackage{graphicx}
\usepackage{textcomp}
\usepackage{xcolor}
\def\BibTeX{{\rm B\kern-.05em{\sc i\kern-.025em b}\kern-.08em
    T\kern-.1667em\lower.7ex\hbox{E}\kern-.125emX}}
\begin{document}

\title{Literature Review of Action Recognition in the Wild\\
}

\author{\IEEEauthorblockN{1\textsuperscript{st} Asket Kaur}
\IEEEauthorblockA{\textit{Computing Science, Multimedia} \\
\textit{University of Alberta}\\
Edmonton, Canada \\
asket@ualberta.ca}
\and
\IEEEauthorblockN{2\textsuperscript{st} Navya Rao}
\IEEEauthorblockA{\textit{Computing Science, Multimedia} \\
\textit{University of Alberta}\\
Edmonton, Canada \\
gururajr@ualberta.ca}
\and
\IEEEauthorblockN{3\textsuperscript{st} Tanya Joon}
\IEEEauthorblockA{\textit{Computing Science, Multimedia} \\
\textit{University of Alberta}\\
Edmonton, Canada \\
tjoon@ualberta.ca}
\and
}

\maketitle

\begin{abstract}
The literature review presented below on Action Recognition in the wild is the in-depth study of Research Papers. Action Recognition problem in the untrimmed videos is a challenging task and most of the papers have tackled this problem using hand-crafted features with shallow learning techniques and sophisticated end-to-end deep learning techniques.
\end{abstract}

\section{Two-Stream Convolutional Networks for Action Recognition in Videos}
\subsection{Authors}
(Karen Simonyan, Andrew Zisserman)
\subsection{Overview}
The authors of this paper investigated architectures that are trained on deep Convolutional Networks for action recognition in video. The challenge in this problem is to capture the complementary information on appearance from still frames and motion between frames. Their aim is to generalise the best performing hand-crafted features within a data-driven framework. Their main contribution is that they were first to propose a two-stream ConvNet architecture which incorporates spatial and temporal networks. They showed that a ConvNet trained on a multi-frame dense optical flow is able to achieve very good performance in spite of limited training data and they demonstrated that multi-task learning applied to two different action classification datasets can be used to increase the amount of training data and improve the performance of both.

Each stream is implemented using a deep ConvNet, softmax scores of which are combined by late fusion. They consider two fusion methods: averaging and training a multi-class linear SVM on stacked L2 Normalised softmax scores as features. The architecture corresponds to CNN-M-2048 and all hidden weight layers use the rectification (ReLU) activation function, maxpooling is performed over 3*3 spatial windows with stride 2. The only difference between the spatial and temporal ConvNet configuration is that they the second normalisation layer from the temporal to reduce memory consumption.

\section{Action VLAD:  Learning spatio-temporal aggregation for action classification}
\subsection{Authors}
(Rohit Girdhar,Deva Ramanan, Abhinav Gupta, Josef Sivic, Bryan Russell)
\subsection{Overview}
In this work, the most notable contribution by the authors is the usage of learnable feature aggregation (VLAD) as compared to normal aggregation using maxpool or avgpool. The aggregation technique is akin to bag of visual words. There are multiple learned anchor-point (say c1, …ck) based vocabulary representing k typical action (or sub-action) related spatiotemporal features. The output from each stream in two stream architecture is encoded in terms of k-space “action words” features - each feature being difference of the output from the corresponding anchor-point for any given spatial or temporal location.

They train all their  networks with a single-layer linear classifier on top of ActionVLAD representation. Throughout, they use K = 64 and a high value for $\alpha$ = 1000.0. Since the output feature dimensionality can be large, they use a dropout of 0.5 over the  representation to avoid overfitting to small action classification datasets. They train the network with cross-entropy loss, where the probabilities are obtained through a softmax. Similar to two stream network, they  decouple ActionVLAD parameters {ck} used to compute the soft assignment and the residual  to simplify learning.

\section{Temporal 3D ConvNets: New Architecture and Transfer Learning for Video Classification}
\subsection{Authors}
(Ali Diba, Mohsen Fayyaz, Vivek Sharma, Amir Hossein Karami, Mohammad Mahdi Arzani, Rahman Yousefzadeh, Luc Van Gool)
\subsection{Overview}
The authors of this paper introduced a new temporal layer that models variable temporal convolution kernel depths. They embedded a new temporal layer in the proposed 3D CNN. They extended the DenseNet architecture - which normally is 2D - with 3D filters and pooling kernels. They called their video convolutional network ‘Temporal 3D ConvNet’ (T3D) and
its new temporal layer ‘Temporal Transition Layer’ (TTL). Their experiments show that T3D outperforms the current state-of-the-art methods on the HMDB51, UCF101 and Kinetics
Datasets. They employed a simple and effective technique to transfer knowledge from a pre-trained 2D CNN to a randomly initialized 3D CNN for a stable weight initialization. This allowed them to significantly reduce the number of training samples for 3D CNNs.

Their goal was to capture short, mid and long term dynamics for a video representation that embodies  more semantic information. Inspired by the idea of GoogleNet they proposed Temporal Transition Layer(TTL). It consists of several 3D convolution kernels, with diverse temporal depths. The TTL output feature maps are densely fed forward to all subsequent layers, and are learned end-to-end. The major contribution of their work is supervision and knowledge transfer between cross architectures from 2D to 3D ConvNets thus avoiding the need to train 3D ConvNets from scratch.

\section{Human Activity Recognition and Prediction: A Survey}
\subsection{Authors}
(Kong, Yu, and  Yun Fu)
\subsection{Overview}
In this paper, the authors talk about the two major problems that are action recognition and action prediction.  In the action recognition task, there are two major types of techniques that can be followed. First is the Shallow Approaches. Shallow Approaches flowchart has two major components one is action representation and the next task is the action classification. Action Representation can be done using Holistic Approaches, Localization Approaches or Trajectory based Approaches. Action classification can be done using Bag-of-the-words approach or sequential approach which uses off the shelf classification algorithms like SVM, kNN. In the Deep Learning Approach, this paper categorises as Space-Time Networks where they talk about 3D ConvNets, C3D, etc., Secondly, Multi-stream Networks where they introduced the famous Two Stream Network and how the development of pooling layer by fusion to improve the interaction between the two layers were proposed and the best pooling layer ActionVLAD is also introduced. Thirdly, Hybrid Networks where they introduced LRCN hybrid of Convnets and LSTMs. 

In the Action Prediction, There are two major categories. One is Short-term prediction and the other is long-term prediction. The authors talk about a paper by Ryoo which proposed integral bag-of-words(IBoW) and dynamic bag-of-words(DBoW) approach, Cao et al built action models by learning feature bases using sparse coding. Li et a  explored long-term accuracy prediction problem.  Lan et al developed hierarchical movements for action prediction. There were deep learning approaches were also discussed in which a new monotonically decreasing loss function in learning LSTMs for action prediction was used. Inspired by that idea the authors implemented an autoencoder to model sequential context information for action prediction. 

\section{Towards Maritime Video surveillance using 4K Videos}
\subsection{Authors}
(V. MARIE, I. BECHAR, and F. BOUCHARA)
\subsection{Overview}
In this paper authors developed  a novel approach to automatic maritime target recognition in the framework of near real-time maritime video-surveillance using super-resolved (i.e.; 4K) videos captured either with a static or with a moving video camera. The proposed approach attempt to leverage both spatial and temporal information for fast and accurate target extraction. The approach is mainly based on the notion of object rigidity and the property that, generally , the maritime objects are texturally richer than the background.

To make it real-time computational efficiency was considered carefully and implemented it in the two levels. Firstly, They proposed a parsimonious implementation of the rigidity measure by assessing the temporal deformation of the SIFT descriptor at key video locations. Second, they developed a multi-resolution scheme for extracting full object zones based both on rigidity analysis and spatial discrimination.

\section{A Heterogeneous image fusion algorithm based on LLC coding}
\subsection{Authors}
(Bing Zhu, Weixin Gao, Xiaomeng Wu, Ruixing Yu)
\subsection{Overview}
The general image fusion algorithm only performs fusion processing on two images. For a large number of hetero-images to be fused, the processing efficiency is poor. Therefore, In this paper the authors choose the method of extracting hetero-images features based on LLC coding and fuses the extracted hetero-images features. This can handle large number of registered infrared and visible light images. The images were encoded and feature of the images was extracted by that codes. It was judged whether the images needed to be merged by the coincidence of the non-zero coding counterpart obtained from comparing the LLC coding of two heterogeneous images. The max-pooling criterion was employed to fuse the feature of images for maximize the complementary information and minimize the
redundant information. Finally, SVM classifier was used to classify and identify the target. 

This method comprises of, construction of codebooks and LLC coding principles, spatial pyramid matching to normalize image features, heterologous image LLC coding and fusion methods. The target recognition rate of images processed by the algorithm were greater than that of unprocessed images

\section{Co-occurrence feature learning for Skeleton Based Action Recognition using regularized Deep LSTM Networks}
\subsection{Authors}
(Wentao Zhu, Cuiling Lan, Junliang Xing, Wenjun Zeng, Yanghao Li, Li Shen, Xiaohui Xie)
\subsection{Overview}
The authors designed a fully connected deep LSTM network for skeleton based action recognition. The proposed model facilitates the automatic learning of feature  co-occurrences from the skeleton joints through the designed regularization. To ensure effective learning of the deep model, they designed an in-depth dropout algorithm for
the LSTM neurons, which performs dropout for the internal gates, cell, and output response of the LSTM neuron. Experimental results demonstrated the state-of-the-art performance
of their model on several datasets.

Their architecture comprised of three bidirectional LSTM layers, two feedforward layers, and a softmax layer that gives the predictions. The proposed full connection architecture enables one to fully exploit the inherent correlations among skeleton joints. In the network, the co-occurrence exploration is applied to the connections prior to the second LSTM layer to learn the co-occurrences of joints/features. LSTM dropout is applied to the last LSTM layer to enable more effective learning. Each LSTM layer uses bidirectional LSTM and they do not explicitly distinguish the forward and backward LSTM neurons. At  each time step, the input to the network is a vector denoting the 3D positions of the skeleton joints in a frame.

\section{Trajectons: Action Recognition Through the Motion Analysis of Tracked Features}
\subsection{Authors}
(Pyry Matikainen, Martial Hebert, Rahul Sukthankar)
\subsection{Overview}
The authors of this paper present a technique for motion based on quantized trajectory
snippets of tracked features. These quantized snippets, or trajectons, rely only on simple feature tracking and are computationally efficient. They demonstrated that with the bag of words framework trajectons can match state of the art results outperforming Histogram of Optical Flow features on the Hollywood actions dataset. Compared to existing motion features their method is able to take advantage of the positive features of each class: the computational efficiency of derivatives and sparse features, the performance of optical flow, and the deep structure of silhouettes. 

Their approach proceeds by the standard bag of words approach. Firstly, the features are tracked over the video using a KLT tracker to produce feature trajectories for a number of features. These trajectories are slightly transformed to produce a number of trajectory snippets for each video. Given a training set of videos, first a dictionary of trajectory words or trajectons is produced by clustering a sample set of trajectory snippets into a specified number(k) of clusters, the centers of which are retained as the trajecton library. Secondly, each video trajectory snippets are assigned a label of the nearest center in the trajecton library, and these labels are accumulated over the video to produce a histogram with k bins. This k length vector is normalized to sum to one for each video, and the training set of histograms along with training class labels is used to train a support vector machine
(SVM) to classify videos into action categories.

\section{Person-centric Multimedia: How research Inspirations from Designing Solutions for Individual Users Benefits the Broader Population}
\subsection{Authors}
(Sethuraman Panchanathan, Ramin Tadayon, Hemanth Venkateswara, and Troy McDaniel)
\subsection{Overview}
The paradigm of Person-centered Multimedia Computing(PCMC) addresses the challenge of Human Centered Multimedia Computing(HCMC) by focusing the design of a system on a single user and challenge, by shifting the focus to the individual. It is proposed that this paradigm can then extend the applicability of multimedia technology from the individual user to the broader population through the application of adaptation and integration.

The authors of this paper have demonstrated the concept of Person-Centric Multimedia computing in Disability Research, Adaptation: Transition to the Broader Public with the example of  Domain Adaptation, Integration: Stealth in Coadaptive Design by giving the example of autonomous training assistant. With all these proof of concept presentation they have concluded that with careful design of the solutions inspired by unique individuals in multimedia need not restrict themselves to that individual instead they can benefit from the broader population. Under this novel paradigm, individual empowerment fostered by a smart, learning, coadapting interface can redene the way we view disability in human-computer interaction. Specically, it can be interpreted in this domain as a spectrum of ability which can be modulated both by the individual and by the technology to create a space in which any individual is able to benet from interacting with this technology.

\section{Skeleton-based action recognition with Convolutional Neural Network}
\subsection{Authors}
(Chao Li, Qiaoyong Zhong, Di Xie, Shiliang Pu)
\subsection{Overview}
Raw skeleton coordinates as well as skeleton motion are fed directly into CNN for label prediction. A novel skeleton transformer module is designed to rearrange and select important skeleton joints automatically. With a simple 7-layer network, they obtained 89.3\% accuracy on validation set of the NTU RGB+D dataset. For action detection in untrimmed videos, they developed a window proposal network to extract temporal segment proposals, which are further classified within the same network. On the recent PKU-MMD dataset, they achieved 93.7\% mAP, surpassing the baseline by a large margin.

For action classification, along with the raw joint coordinates, motion of skeleton joints from two consecutive frames are fed as an extra input to the network and they proposed a novel skeletal transformer module with that module, the network is able to automatically learn a better ordering of joints as well as new joints that are more informative than arbitrarily given ones. And to deal with multiple people they used maxout to merge features from skeletons of different individuals.  For action detection, a sequence of skeletal data as a T*N*3 image, it is able to adapt object detection methods to the task. They used Faster R-CNN where they had changed window proposal network in particular 2D anchors are flattened to 1D anchors.

\section{Spatial Temporal Graph Convolutional Networks for Skeleton-Based Action Recognition}
\subsection{Authors}
(Sijie Yan, Yuanjun Xiong, Dahua Lin)
\subsection{Overview}
In this paper by extending graph neural networks to a spatial-temporal graph model called Spatial-Temporal Graph Convolutional Networks, they design a generic representation of skeleton sequences for action recognition. This model is formulated on top of a skeleton graph sequence, where each node corresponds to a human body joint.This creates multiple layers of spatial temporal graph convolution, allowing data to be combined along both the spatial and temporal dimensions. Their proposed model achieves superior performance compared to previous approaches using hand-crafted parts or traversal rules on two large-scale data sets for skeleton-based action recognition, with substantially less effort in manual development. In dynamic skeleton sequences, ST-GCN can capture movement information that is complementary to RGB modality. The combination of skeleton-based model and frame-based model further improves action recognition efficiency.

\section{Human Action Recognition by Representing 3D Skeletons as Points in a Lie Group}
\subsection{Authors}
(Raviteja Vemulapalli, Felipe Arrate and Rama Chellappa)
\subsection{Overview}
In this paper,they propose  action recognition for a new body part-based skeletal representation. Inspired by the fact that the relative geometry between different parts of the body provides a more accurate explanation than their absolute positions for human actions, they specifically model the relative 3D geometry in their skeletal representation of different parts of the body. The relative geometry can be represented using the rotation and translation required to take one body part to  the position and orientation of the other. Rigid body rotations and translations in 3D space  are members of the special Euclidean group. Joint positions and relative joint positions when used with the temporal modeling and classification  produce results better than the state-of-the-art reported on UTKinect-Action and Florence3D-Action datasets. This suggests that the combination of DTW, FTP and linear SVM is well-suited for skeleton-based action classification. They experimentally showed that the proposed representation performs better than many existing skeletal representations on three different action datasets. In their work, they used the relative geometry between all pairs of body parts. But, each action is usually characterized by the interactions of a specific set of body parts. Hence, they are planning to explore various strategies to automatically  identify the set of body parts that differentiates a given action from the rest. In this paper, they focused only on actions  performed by a single person and are planning to extend this representation to model multi-person interactions.

\section{Spatio-Temporal Graph Routing for Skeleton-Based Action Recognition}
\subsection{Authors}
(Bin Li, Xi Li,Zhongfei Zhang, Fei Wu)
\subsection{Overview}
In this paper, they propose a new spatio-temporal graph routing scheme for skeleton-based action recognition that learns the intrinsic high-order connectivity relationships for physically separate skeleton joints in an adaptive way. The SGR seeks to discover the connectivity relationships between the joints based on the spatial dimension clustering of subgroups, while the Temporal GR investigates the structural  information by measuring the correlation degrees between temporal joint node trajectories. They put forward a novel Spatio-Temporal Graph Routing scheme to model the semantic connections among the joints in a disentangled way. Rather than using fixed human skeleton, two sub-networks are responsible to capture both spatial and temporal dependencies between each two nodes, serving as routers for all nodes. Their STGR-GCN model, with simple spatio-temporal routing method, presents better results compared with  vanilla ST-GCN and further achieves state-of-the-art, implying the effectiveness of dynamic routing scheme among  graph convolution layers.

\section{A Flexible Method for Time-of-Flight Camera Calibration Using Random Forest}
\subsection{Authors}
(Chi Xu and Cheng Li)
\subsection{Overview}
In this work, they propose a learning-based approach using random forest to calibrate the ToF depth images. They learn from empirical observations of the complex geometric distortion model of the ToF camera without prior assumptions on the model. They aim at directly calibrating the processed ToF depth images as provided by the manufacturer. The calibrated depth image is expected to preserve local geometric and topological properties of the scene objects. As the distortion pattern of ToF camera is complex and irregular, random forest is a powerful tool to discover the geometric distortion model. Their calibration procedure is fast and simple, so the camera can be mobile and easily used in different settings. Their proposed method is flexible, and the calibration process needs only a ToF camera and a standard chessboard. The training data is collected by observing the chessboard from different viewpoints and distances, and it can be easily performed by a non-expert user.

\section{Detection-based Online Multi-Target Tracking via Adaptive Subspace Learning}
\subsection{Authors}
(Jyoti Nigam, Krishan Sharma, and Renu M Rameshan)
\subsection{Overview}
In this work, they address the issue of multi-target tracking subject to varying number of targets, variation throughout appearance due to camera movement, as well as targets along with variation in pose and illumination. They suggest an adaptive MTT learning approach for subspace learning, integrating selected new observations into the learned subspace. This approach prevents unlimited subspace development, which makes our tracker computationally better. Addressing merging and splitting and occlusion problems by using reconstruction errors and not updating the occluded person's corresponding subspace, respectively. The data sets used are from the benchmark of MOTChallenge published as 2D MOT15, MOT16 and MOT17. The algorithm begins with the formation of initial subspaces followed by data association i.e. matching detections with existing trajectories. The adaptive subspace learning helps the algorithm in keeping track of changes in target appearance which may occur throughout the length of the video. Their main assumption is that a moving target forms a low dimensional subspace in some feature space.

\section{A Deep Learning Approach to Predict Crowd Behavior Based on Emotion}
\subsection{Authors}
(Elizabeth B Varghese and Sabu M Thampi)

\subsection{Overview}
In this work, a broad representation system for classifying crowd behaviour based on crowd emotions is proposed. Instead of using appearance or movement features or other low-level features to model crowd behaviour patterns, they use deep spatio-temporal features that learn to distinguish crowd emotions using a 3D Convolutional Neural Network(3DCNN). A multiclass Support Vector Machine (SVM) to predict multiple heterogeneous crowd behaviors. The features from the 3DCNN encapsulate information from video frames related to different emotions. The emotions are mapped to their corresponding crowd behaviors using SVM and their method is able to predict six different behavioral classes such as fight, normal, abnormal object, panic, cheerful and congested which is more than most of the state of the art methods. They use the Motion Emotion Dataset (MED) to extract the emotional features as the dataset has emotionally annotated video frames that are very useful for extracting features using 3DCNN. The proposed deep framework with the multiclass model outperforms the methods that use low-level features to classify crowd behavior and abnormal activity classification. Their work showed that emotions extracted from spatio-temporal features are outstanding in bridging the linguistic gap between low-level descriptors and high-level discrete crowd behaviors.

\section{Trajectory-Based Modeling of Human Actions with Motion Reference Points}
\subsection{Authors}
(Yu-Gang Jiang, Qi Dai, Xiangyang Xue, Wei Liu, Chong-Wah Ngo)

\subsection{Overview}
This paper proposes an approach to model the motion relationships among moving objects and the background. They introduce two kinds of reference points to characterize complex motions in the unconstrained videos, in order to alleviate the effect incurred by camera movement. Tracking of local frame patches is firstly performed to capture the motion of the local patches. Next, local frame patch monitoring is conducted to capture the regional patch movement. They use a simple clustering approach with the trajectories to define the scene's dominant motion, which is used as a global motion reference point to calibrate that trajectory's motion. Each trajectory is used as a local motion reference point for motion characterization, thus their representation is naturally robust to camera motion as it only counts the relative motion between trajectories. The object relationships are encoded by the motion patterns among pairwise trajectory codewords, so that accurate object boundary detection or foreground-background separation is avoided.Extensive experiments on three challenging action recognition benchmarks have shown that the proposed approach offers very competitive results. Their approach already is complementary to the standard bag-of-features.

\section{Activity recognition using the velocity histories of tracked keypoints}
\subsection{Authors}
(Ross Messing, Chris Pal, Henry Kautz)

\subsection{Overview}
In this paper, they present a psychophysical performance-inspired activity recognition function. This feature is based on tracked keypoint velocity history. Using this feature, they present a generative mixture model for video sequences and show that it performs well on the KTH event recognition dataset in comparison with local spatiotemporal features.
Their background function performs well on complicated events in high-resolution video sequences. They also demonstrate how to expand the velocity history function, both with a more complex latent velocity model, and by combining the velocity history feature with other useful information such as size, position, and semantic high-level information.
The velocity history of tracked keypoints is a useful feature for activity recognition, particularly when increased with additional information. Through integrating information at the feature level outside a regional spatio-temporal patch, a bag-of-functions model allows for a large amount of non-local structure.

\section{Action Recognition with Improved Trajectories}
\subsection{Authors}
(Heng Wang and Cordelia Schmid)

\subsection{Overview}
By explicitly estimating camera motion, this paper improves dense trajectories. They show that performance can be improved significantly by removing background trajectories and warping optical flow with a robustly estimated homography approximating camera motion. During camera movement estimation, potentially inconsistent matches can be removed using a state-of - the-art human detector. A systematic analysis of four complex datasets reveals the feasibility of the suggested solution and sets new quality limits.

\section{Action Recognition by Dense Trajectories}
\subsection{Authors}
(Heng Wang, Alexander Kläser, Cordelia Schmid, Liu Cheng-Lin)

\subsection{Overview}
In this paper they propose an approach to describe videos by dense trajectories.
They sample dense points from each frame and monitor them from a dense optical flow field based on displacement data. Their trajectories are robust to strong irregular motions as well as shot boundaries by using a good optical flow algorithm. A novel descriptor based on histograms of motion boundary that is resilient to camera movement is implemented in their approach. This descriptor consistently outperforms other state-of-the-art descriptors, in particular in uncontrolled realistic videos. 

They evaluate the video description in the context of action classification with a bag-of-features approach. They capture the motion information in the videos efficiently and show improved performance over state-of-the-art approaches for action classification. Their descriptors combine trajectory shape, appearance, and motion information.Their approach also introduces an efficient solution to remove camera motion by computing motion boundaries descriptors along the dense trajectories. This successfully segments the relevant motion from background motion, and outperforms previous video stabilization methods.

\section{Person Authentication by Air-writing Using 3D Sensor and Time Order Stroke Context}
\subsection{Authors}
( Lee-Wen Chiu, Jun-Wei Hsieh, Chin-Rong Lai, Hui-Fen Chiang , Shyi-Chy Cheng, and Kuo-Chin Fan )

\subsection{Overview}
The signature authentication system can be off-line based or on-line based. Off-line based systems can be hacked easily as compared to on-line based systems because it does not capture the writing speed information of signature and thus a comparison between two signatures is majorly based on their shape similarity. As the writing speed of signature can not be imitated easily therefore, They proposed air-writing signature authentication system without using any pen-starting-lift-signal. The previous method used Fourier descriptors and moment features to match the writing signatures for user authentication which was not robust to noise and redundant strokes. Therefore, they used reverse-time - ordered stroke representation to filter out the captured air writing trajectories. This backward representation of the trajectory easily filtered out the trajectory and simplified the matching process as a path-finding problem. The method used the weighting scheme, to solve the pathfinding problem in real-time via a dynamic time warping technique. Since the signature of a particular user can vary, therefore they used agglomerative hierarchical clustering scheme to cluster user signature into different subclasses and used average within-class distance to determine whether a user’s signature is passed or not. The method adopted, the dataset including 1326 signatures and achieved authentication accuracy of more than 93.5\% with no starting gesture required.

\section{Research on path planning method of an unmanned vehicle under urban road environment}
\subsection{Authors}
(Yu Ruixing and Zhu Bing and Cao Meng and Zhao Xiao and Wang Jiawen)

\subsection{Overview}
The authors of this paper developed a composite path planning algorithm for an unmanned vehicle driving in a complex urban environment in China by combining the A* and Stochastic Fractal Search(SFS) algorithm. As urban roads in China are very complicated and have a mixture of people and various types of vehicles running on roads with several road intersections. Therefore, rather than considering the vehicle as a particle, They used the front-wheel-drive model as a more feasible vehicle model. The previous method used only SFS algorithm for path planning, which worked very well in complex environments but the method took several hours to find a collision-free path hence, reduced applicability in real-time applications. Therefore, this method compared combined the dynamic nature of the A* algorithm and ability of the SFS algorithm to handle multiple constrained optimization problems. They used A* algorithm to obtain the trajectory of the locus of centre point P of the rear wheel of the vehicle and finally used SFS to calculate collision-free path with vehicle’s state information. The paper further compared the A* algorithm with APF and RRT algorithm and found that A* algorithm took the least time of 0.2381 seconds for path computation. The paper hence, concluded that using compound algorithm is more efficient than using only the SFS algorithm for path planning of unmanned vehicle.

\section{Synthetic Vision Assisted Real-time Runway Detection for Infrared Aerial Images}
\subsection{Authors}
(Changjiang Liu, Irene Cheng and Anup Basu)

\subsection{Overview}
In aviation, approaching and landing under low visibility and night vision conditions is the major safety issue. This paper focused only on the runway analysis as it is the prominent region of detection. To overcome the difficulty in differentiating the touch zone and actual runway in infrared images. The method  generalised Otsu's thresholding method for trichotomizing the area that overlapped a virtual runway and thus, obtained the initial curve developed by level set methods. In addition, it proposed a scheme based on ROI to speed up the level set methods. This method produced more accurate runway outline. The proposed method significantly improved the time performance and the strategy was applicable for real time processing without any manual intervention.

\section{Optimized Skeleton-based Action Recognition via Sparsified Graph Regression}
\subsection{Authors}
(Xiang Gao, Wei Hu, Jiaxiang Tang, Jiaying Liu, Zongming Guo)

\subsection{Overview}
To overcome the challenge of irregular skeleton joints and graph construction faced in previous methods using RNN, CNN and GCN(graph convolutional networks), They naturally represented the skeletons on graph where they represented each joint as a vertex and the relation between the joints as edges. For recognizing actions based on skeleton, They proposed a graph regression based GCN (GR-GCN) which used multiple observations to statistically learnt the  common sparsified representation of graphs. They obtained efficient and optimized graph representation connecting not only each joint with its neighboring joints but connecting the relevant joints from previous frames to the subsequent frames by providing spatio-temporal modeling and optimizing the graph structure over the consecutive frames. The optimized graph was then fed into GCN for learning features, where they used high order and fast Chebyshev spectral graph convolution approximation. The proposed GR-GCN method achieved the state-of-the-art performance on the UT-Kinect,SYSU 3D and NTU RGB+D datasets.

 \section{Rolling Rotations for Recognizing Human Actions from 3D Skeletal Data}
\subsection{Authors}
( Raviteja Vemulapalli, Rama Chellappa)

\subsection{Overview}
The authors proposed the human-action recognition method using rolling-map from 3D skeleton data. As 3D geometry between the various body parts provide more meaningful description of the human skeleton in action recognition as compared to absolute locations. Therefore, They used 3D rotation between different parts of the body to represent the different skeletons. Then they computed nominal curve using DTW and during this step they used the Lie algebra distance to speed up the computation. The proposed method then performed rolling and unwrapping operation followed by Linear SVM classification for human action recognition. They evaluated the proposed method on three different action datasets, which showed that the proposed method outperforms various skeleton based state-of-the-art action recognition approaches.

\section{Super Normal Vector for Human Activity Recognition with Depth Cameras}
\subsection{Authors}
( Xiaodong Yang and YingLi Tian,)

\subsection{Overview}
With the development of Microsoft Kinect and various other low cost depth cameras, number of researches started using depth cameras for capturing video sequences for recognizing human activities. They proposed an approach based on the collection of neighboring hypersurface normals that is low-level polynormals from the local spatiotemporal depth volume which are then aggregated by SNV(Super Normal Vector) in each adaptive spatiotemporal cell. The sequence of feature vectors from all the spatiotemporal cells constituted the final depth sequence representation. They used linear classifier for classifying the actions. The proposed method was then experimented on 4 different datasets that are MSR Action3D, MSR Gesture3D , MSR Action Pair3D and MSR Daily Activity3D dataset on which it showed 93.45\%, 94.74\%, 100\% and 86.25\% accuracy respectively.

\section{A Novel Hierarchical Framework for Human Action Recognition}
\subsection{Authors}
(Hongzhao Chen, Guijin Wang, Jing-Hao Xue , Li He)

\subsection{Overview}
To tackle the problems of high intra-class variation, the velocity of motion and high computational costs involved in action recognition. Chen et al.proposed the two-level hierarchical structure based on skeletons. A part based clustering feature vector was implemented in the first layer to evaluate the most important joints and group them into an initial classification. The task of recognition was further divided into a number of simpler and smaller tasks performed within a single cluster to resolve the high intraclass variation as there are multiple sequences of the same actions are clustered into different clusters. In the second layer, only the related joints in different clusters were used for extracting features, thereby improving the validity of the features and the computational costs. The proposed method was experimented with the two different datasets i.e MSR Action3D and UTKinect Action dataset which showed around 95-96\% and 95.96\% accuracy respectively for the datasets.

\section{A Depth Video-based Human Detection and Activity Recognition using Multi-features and Embedded Hidden Markov Models for Health Care Monitoring Systems}
\subsection{Authors}
(Ahmad Jalal , Shaharyar Kamal and Daijin Kim)

\subsection{Overview}
The authors considered depth video and proposed a method using multi-features and embedded HMMs(Hidden Markov Model, for extracting human 3D silhouettes and spatiotemporal joints for their compact and adequate Human action recognition task data. The proposed HAR system, initially analysed  depth maps using the method of temporal motion recognition to segment human silhouettes from a noisy background and then measured the depth silhouette area for each event to track human movements within the scene. To explore changes in the gradient orientation, differentiation of intensity, local movement of specific body parts and temporal variation, They combined various representative features. These features were then processed by their respective class dynamics and afterwards it learned, modelled, trained and recognised with the specific embedded Hidden Markov Model with active feature values. The proposed method was experimented on three different datasets i.e Online self-annotated dataset and MSR Daily Activity 3D and MSR Action3D dataset in which the method showed the accuracy of 71.6\%, 92.2\% and 93.1\% respectively.

\section{3-D Human Action Recognition by Shape Analysis of Motion Trajectories on Riemannian Manifold}
\subsection{Authors}
(Maxime Devanne, Hazem Wannous, Stefano Berretti,Pietro Pala, Mohamed Daoudi and Alberto Del Bimbo)

\subsection{Overview}
This paper proposed a new framework for extracting the dense representation of a human action captured by a depth sensor for accurate recognition of action. The proposed approach developed human skeleton model that acquired the data and represented the joints as the 3-D coordinates and the change in the coordinates over time as a trajectory in the action space. The system simultaneously captured both the shape and movement of the human body. Then for action recognition, similarity between the shapes of trajectories is computed using Riemannian manifold. Finally the classification was performed on this manifold using k-nearest neighbours,  in the open curve shape space with the help of Riemannian geometry. This method was experimented on four different benchmarks and showed 90\% accuracy. The proposed method outperformed various state-of-art approaches.

\section{Detecting Attention in Pivotal Response Treatment Video Probes}
\subsection{Authors}
(Corey DC Heath, Hemanth Venkateswara, Troy McDaniel and Sethuraman Panchanathan )

\subsection{Overview}
Pivotal Response Training (PRT) is a process for applying the Applied Behavior Analysis (ABA) scientific principles to teach students with autism, social communication skills and behavioural skills in a naturalistic learning environment. Video probes are an important part of assessing people learning PRT. The manual process is constrained by the expense of obtaining relevant data from behavioural analysts. Therefore, this paper proposed a machine learning technique which can be used in classifying the video probes. The primary focus of this paper is to examine how video processing can automatically infer attention. To achieve this, they created a dataset using PRT sessions video probes and used it to train the models of machine learning. The paper concluded that the complexity in these videos gave the learning of an intelligence feedback system a significant set of challenges.

\section{CONCLUSION}

This paper reviewed various video analysis approaches used in different applications like human action recognition, runway detection, etc and provides a brief summary of the methodology used in each of the reviewed paper. The paper also mentions the various datasets on which the proposed method was experimented and provides the brief knowledge about their accuracy.

\end{document}